\documentclass[11pt]{article}

\usepackage[preprint]{acl}

\usepackage{times}
\usepackage{latexsym}
\usepackage[T1]{fontenc}
\usepackage[utf8]{inputenc}
\usepackage{microtype}
\usepackage{inconsolata}
\usepackage{graphicx}

\usepackage{subcaption}
\usepackage{booktabs}
\usepackage{multirow}

\usepackage{amsmath}
\usepackage{amssymb}
\usepackage{amsfonts}
\usepackage{mathtools}
\usepackage{amsthm}
\usepackage{bm}

\usepackage{algorithm}
\usepackage{algpseudocode}

\usepackage{xcolor}
\usepackage{enumitem}
\usepackage[capitalize,noabbrev]{cleveref}

\theoremstyle{plain}

\theoremstyle{definition}

\theoremstyle{remark}

\title{Elastic Horizon: Discovering the Effective Interaction Frontier \\
       in Agentic Reinforcement Learning}

\author{
  \textbf{Gangyi Zhang\textsuperscript{1}\thanks{Equal contribution.}},
  \textbf{Junjie Meng\textsuperscript{2}\footnotemark[1]},
  \textbf{Letian Zhang\textsuperscript{3}},
  \textbf{Wei Wu\textsuperscript{2}},
  \textbf{Yang Zheng\textsuperscript{3}},
\\
  \textbf{Dong Wang\textsuperscript{1}\thanks{Corresponding authors.}},
  \textbf{Yang Liu\textsuperscript{1}},
  \textbf{Guanjun Jiang\textsuperscript{1}},
  \textbf{Chongming Gao\textsuperscript{2}\footnotemark[2]}
\\
\\
  \textsuperscript{1}Qwen Business Unit of Alibaba,
  \textsuperscript{2}University of Science and Technology of China,
\\
  \textsuperscript{3}Independent Researcher
\\
\\
  \normalfont\small\ttfamily gangyi.cn@gmail.com, mengjre@gmail.com
\\
  \normalfont\small\ttfamily wd443495@alibaba-inc.com, chongming.gao@gmail.com
}

\begin{document}
\maketitle


\begin{abstract}
Scaling the interaction horizon---the maximum number of environment interactions per episode---improves LLM agents on long-horizon tasks, and curriculum-based methods that progressively expand the horizon outperform fixed-horizon alternatives.
However, existing schedules are \emph{open-loop}: they monotonically increase the horizon until a manually specified maximum, with no mechanism to detect when further expansion stops helping.
We propose the \emph{effective interaction frontier hypothesis}: a dynamic boundary beyond which additional interactions yield diminishing returns while cost grows linearly. We then introduce \textbf{Elastic Horizon}, a closed-loop controller that tracks this boundary via the 90th percentile of successful trajectory lengths.
On AppWorld and BFCL, fixed-horizon sweeps reveal clear saturation plateaus; Elastic Horizon stabilizes the horizon inside the saturation band from both under- and over-capacity initializations, attains the best success rates across 7B and 14B backbones, and saves up to 25\% of per-step trajectory tokens.
Our work shifts the paradigm from \emph{how to scale} interaction horizons to \emph{when to stop scaling}.
\end{abstract}


\section{Introduction}
\label{sec:introduction}

In partially observable environments, LLM agents must gather information through multi-turn interactions to reduce state uncertainty and complete tasks~\citep{yao2023react, shinn2023reflexion}. 
Recent work demonstrates that \emph{scaling the interaction horizon}---the maximum number of environment interactions per episode---substantially improves performance on long-horizon agentic tasks. 
\citet{xi2025agentgymrl} introduce ScalingInter-RL, a curriculum that linearly increases the horizon during training, matching or surpassing commercial models on 27 tasks across diverse environments. 
\citet{shen2025tti} show that a multiplicative schedule outperforms both fixed-horizon training and additive schedules on web navigation benchmarks.

Despite their success, these approaches share a fundamental characteristic: they are \emph{open-loop monotonically increasing} schedules. 
The horizon grows according to a predetermined function of training steps, continuing until reaching a manually specified maximum $K_{\max}$, with no mechanism to assess whether further expansion remains beneficial. 
This design implicitly assumes that performance improves monotonically with horizon length. We challenge that assumption in this work.

\begin{figure*}[htbp]
    \centering
    \begin{subfigure}{\columnwidth}
        \centering
        \includegraphics[width=\textwidth]{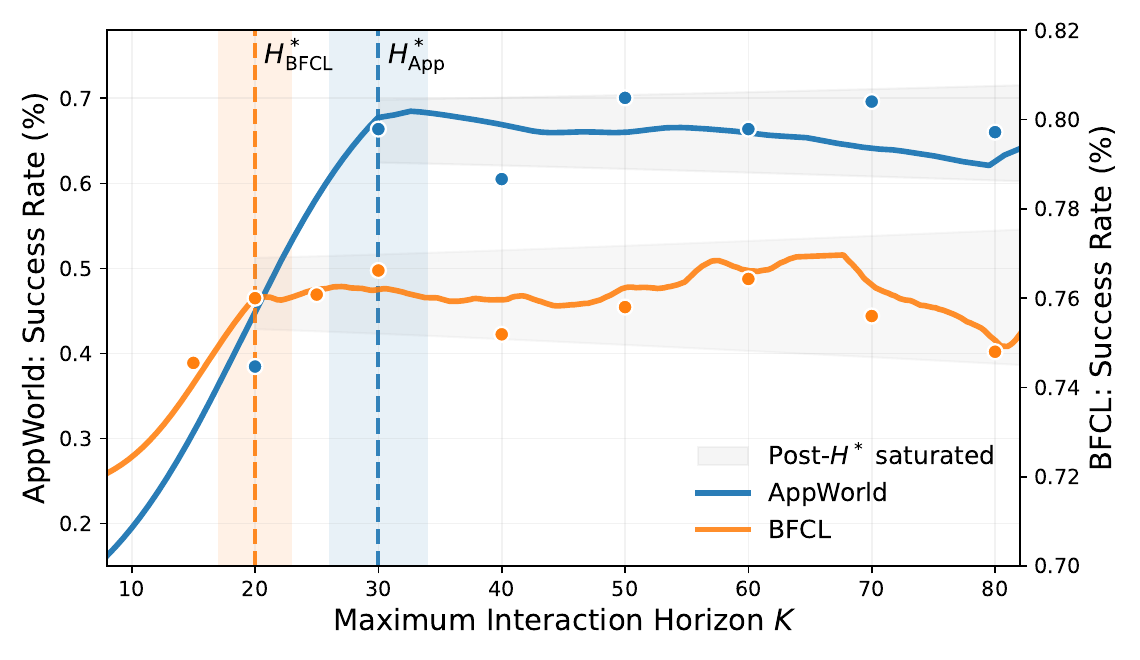}
        \caption{Fixed-horizon training on AppWorld: success rate improves with horizon $K$ until reaching a plateau around the effective frontier $H^*$ (shaded region), beyond which additional budget yields diminishing returns while cost grows linearly.}
        \label{fig:sub1}
    \end{subfigure}
    \hfill%
    \begin{subfigure}{\columnwidth}
        \centering
        \includegraphics[width=\textwidth]{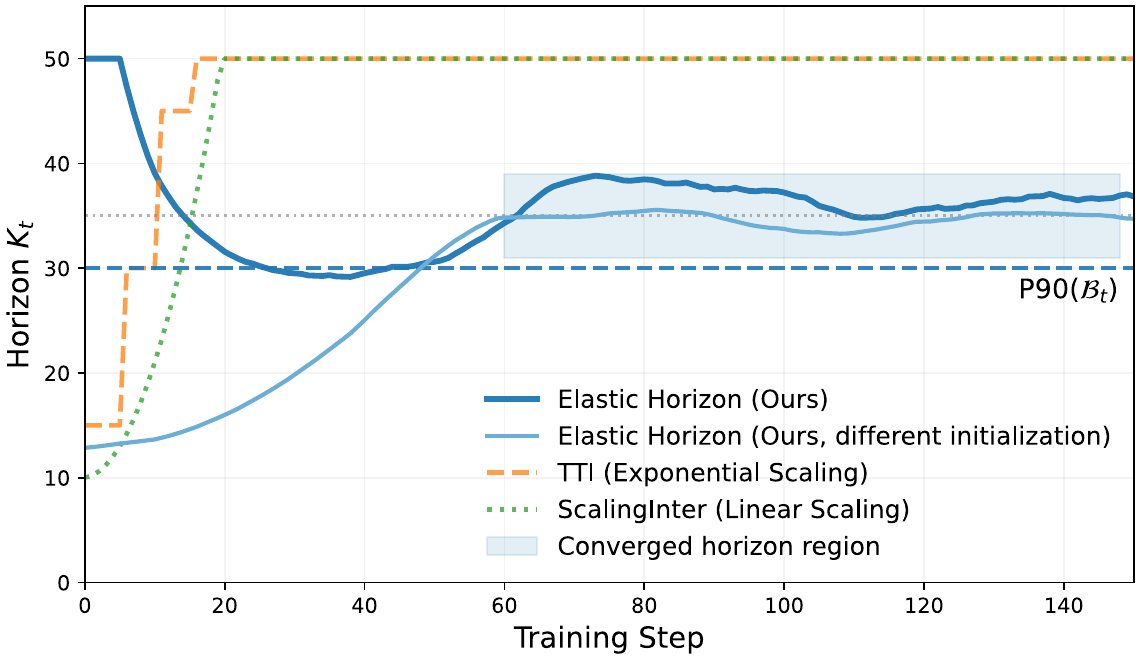}
        \caption{Horizon evolution during training: open-loop schedules (ScalingInter, TTI) increase monotonically toward $K_{\max}=50$, whereas Elastic Horizon stabilizes the horizon inside the saturation band from both low ($K_0=10$) and high ($K_0=50$) initializations.}
        \label{fig:sub2}
    \end{subfigure}
    
    \caption{Interaction horizon saturation and Elastic Horizon.}
    \label{fig:total_figure}
\end{figure*}

We present evidence that horizon scaling exhibits \emph{diminishing marginal returns}. 
Figure~\ref{fig:sub1} shows results from fixed-horizon sweep experiments on AppWorld~\citep{trivedi2024appworld}: success rate increases with the interaction budget $K$ in the under-capacity regime, then \emph{plateaus} around a threshold we term the \emph{effective interaction frontier} $H^*$. 
Beyond this point, additional budget produces no systematic improvement while computational cost---proportional to trajectory length---continues to grow linearly.

This saturation phenomenon has been observed independently in prior work. 
\citet{xi2025agentgymrl} report that RL with a large interaction budget ``quickly collapses,'' and that beginning with a large number of interaction turns ``leads the model into redundant reasoning and unproductive actions.''
\citet{liu2025bats} find that a ReAct agent ``saturates at a budget of 100'' tool calls and ``fails to utilize additional tool budget, reaching a performance ceiling.''
\citet{shen2025tti} observe that a fixed long horizon ($h=30$) leads the agent to expend steps unproductively early in training, degrading performance relative to a shorter horizon.

These observations converge on a common insight: there exists an \emph{effective interaction frontier} $H^*$, jointly determined by task complexity and agent capability, beyond which additional interaction steps yield diminishing returns.
This motivates a fundamental question: \textbf{can we automatically detect and adapt to this frontier?}

We answer affirmatively with \textbf{Elastic Horizon}, a closed-loop horizon controller that automatically discovers and tracks $H^*$ during training.\footnote{Code and training configurations are available at \url{https://github.com/junjie-meng/ElasticHorizon}.}
Unlike open-loop schedules that increase the horizon based on training progress, Elastic Horizon adjusts based on the agent's \emph{demonstrated capability}, specifically the lengths of trajectories that successfully complete tasks.
It estimates the agent's competence boundary using the 90th percentile of successful trajectory lengths, a statistic that captures near-maximal capability while filtering outlier noise from inefficient explorations.

The controller is \emph{bidirectional}: expanding when the agent demonstrates capability growth, and contracting when approaching the task's complexity ceiling (Figure~\ref{fig:sub2}). 
This enables automatic convergence to $H^*$ from arbitrary initializations. Open-loop schedules can only increase the horizon, so they cannot converge from an over-capacity start.

Elastic Horizon complements test-time budget-aware methods like BATS~\citep{liu2025bats}. 
While BATS optimizes deployment performance by injecting budget awareness into prompts, Elastic Horizon optimizes training efficiency by adapting the horizon to demonstrated capability.
The two approaches target different phases and can be combined.

Our contributions are:
\begin{itemize}[leftmargin=*, itemsep=2pt, topsep=2pt]
    \item \textbf{Empirical discovery.} We document interaction horizon saturation on AppWorld~\citep{trivedi2024appworld} and BFCL v3~\citep{patil2025bfcl}, revealing clear plateaus beyond a task-dependent threshold $H^*$ (Section~\ref{sec:exp_saturation}).

    \item \textbf{Methodology.} We propose Elastic Horizon, the first closed-loop horizon controller for agentic RL. It adapts via percentile-based boundary estimation with EMA smoothing, requiring no manual schedule (Section~\ref{sec:method}).

    \item \textbf{Mechanism validation.} Elastic Horizon stabilizes the horizon inside the saturation band from both high and low initializations, validating its role as an automatic frontier detector (Section~\ref{sec:exp_convergence}).

    \item \textbf{Practical impact.} Elastic Horizon attains the best results across both 7B and 14B backbones while reducing per-step trajectory tokens by up to 25\% on AppWorld (Section~\ref{sec:exp_main}).
\end{itemize}

\noindent Together, these contributions shift the focus from \emph{how to scale} interaction horizons to \emph{when to stop scaling}.

\section{Related Work}
\label{sec:related}

\subsection{Multi-Turn Agent Training and Horizon Scaling}
\label{sec:related_horizon}

Training LLM agents for long-horizon tasks presents challenges in stability and sample efficiency. 
\citet{xi2025agentgymrl} introduce AgentGym-RL and propose ScalingInter-RL---a curriculum that progressively extends interaction horizons during training---finding that fixed large horizons cause training instability while gradual expansion improves performance. 
\citet{shen2025tti} demonstrate that scaling interaction steps outperforms scaling per-step reasoning tokens under fixed compute budgets, advocating a multiplicative horizon schedule.
Both methods employ \emph{open-loop} schedules where the horizon increases according to predetermined functions of training steps. 
Elastic Horizon differs by using \emph{closed-loop} control: the horizon adapts based on observed trajectory statistics, enabling automatic convergence to the task's effective frontier without manual schedule design.

Several works address long-horizon challenges through complementary approaches. 
\citet{zhou2024archer} propose ArCHer, combining hierarchical off-policy learning with on-policy token-level optimization.
\citet{wang2025ragen} document the ``Echo Trap'' phenomenon and propose variance-based filtering for stabilization.
\citet{qi2025webrl} introduce self-evolving task curricula, and \citet{chen2025loop} develop memory-efficient PPO variants.
These methods address orthogonal aspects of long-horizon training and can be combined with Elastic Horizon's adaptive horizon control.

\subsection{Budget-Aware Agents}
\label{sec:related_budget}

A parallel thread addresses resource allocation at inference time. 
\citet{liu2025bats} propose BATS, which injects remaining budget information into prompts and adapts planning strategies based on available resources.
\citet{snell2024scaling} establish that compute-optimal test-time strategies should adaptively allocate resources based on prompt difficulty.
\citet{paglieri2025planning} learn \emph{when} to invoke planning versus direct action.
\citet{han2025tale} dynamically estimate token budgets based on reasoning complexity.

These methods focus on \emph{test-time} resource optimization with explicit budget signals. 
Elastic Horizon operates during \emph{training} using implicit capability signals (trajectory length distributions). 
The approaches are complementary: Elastic Horizon optimizes training efficiency, while methods like BATS optimize deployment performance.

\subsection{Curriculum Learning}
\label{sec:related_curriculum}

Curriculum learning~\citep{bengio2009curriculum} demonstrates that presenting examples from easy to hard improves generalization. 
Self-paced learning~\citep{kumar2010selfpaced} extends this by determining difficulty from the learner's current loss, making the curriculum data-driven.
\citet{jiang2015selfpaced} unify these approaches, combining prior difficulty knowledge with dynamic learner feedback.
In reinforcement learning, curriculum methods have been applied to task selection~\citep{narvekar2020curriculum}, goal generation~\citep{florensa2018automatic}, and start state distributions~\citep{florensa2017reverse}.

Elastic Horizon contributes a novel competence signal---the upper percentile of successful trajectory lengths---applied to episode horizon length, a previously underexplored curriculum dimension.
This signal directly measures demonstrated capability to complete long-horizon tasks, making it particularly suited for agentic RL where trajectory length correlates with task complexity.

\subsection{Scaling Laws}
\label{sec:related_scaling}

Scaling laws research establishes that optimal resource allocation depends on model capacity and task demands.
\citet{kaplan2020scaling} derive power-law relationships between performance and compute.
\citet{hoffmann2022chinchilla} demonstrate that model size and training tokens should scale proportionally for compute-optimal training.
\citet{hilton2023scaling} extend these findings to RL, showing that varying the task's horizon length changes the coefficient but not the exponent of the compute-optimal scaling relationship.
These principles support Elastic Horizon's premise: interaction horizons should adapt to agent competence rather than follow fixed maximal allocations.

Table~\ref{tab:positioning} summarizes Elastic Horizon's positioning among related methods.


\begin{table}[t]
\centering
\caption{Positioning of Elastic Horizon.$^\dagger$}
\label{tab:positioning}
\footnotesize
\setlength{\tabcolsep}{3pt}
\begin{tabular}{lccc}
\toprule
\textbf{Method} & \textbf{Stage} & \textbf{Control} & \textbf{Signal} \\
\midrule
ScalingInter-RL & Train & Open-loop & Training step \\
TTI & Train & Open-loop & Training step \\
BATS & Test & Closed-loop & Budget count \\
Self-Paced & Train & Closed-loop & Loss value \\
\textbf{Elastic (Ours)} & \textbf{Train} & \textbf{Closed-loop} & \textbf{Traj.\ length} \\
\bottomrule
\end{tabular}
\vspace{1mm}
{\footnotesize $^\dagger$Elastic Horizon uses the 90th percentile of successful trajectory lengths; see Section~\ref{sec:method_estimation}.}
\end{table}

\section{Preliminaries}
\label{sec:preliminaries}

\subsection{Multi-Turn Agentic Tasks as POMDPs}
\label{sec:prelim_pomdp}

We formalize multi-turn agentic tasks as episodic decision problems with sparse terminal rewards. Following standard practice in agentic RL~\citep{xi2025agentgymrl, wang2025ragen}, we adopt a simplified Partially Observable Markov Decision Process (POMDP) formulation:
\begin{equation}
    \mathcal{M} = (\mathcal{U}, \mathcal{S}, \mathcal{A}, \mathcal{O}, T, R),
    \label{eq:pomdp}
\end{equation}
where $\mathcal{U}$ is the instruction space specifying task goals, $\mathcal{S}$ is the state space (only partially observable to the agent), $\mathcal{A}$ is the action space (e.g., tool calls, API requests), $\mathcal{O}$ is the observation space (e.g., API responses, environment feedback), $T: \mathcal{S} \times \mathcal{A} \to \Delta(\mathcal{S})$ is the state transition function, and $R: \mathcal{S} \to \{0, 1\}$ is a sparse binary terminal reward indicating task success.

The agent's policy $\pi_\theta$ is parameterized by a large language model with parameters $\theta$. Given an instruction $u \in \mathcal{U}$ and the interaction history $h_k = (a_1, o_1, \ldots, a_{k-1}, o_{k-1})$, the policy outputs a distribution over actions: $\pi_\theta(a_k | u, h_k)$. We denote a complete trajectory as $\tau = (u, a_1, o_1, \ldots, a_L, o_L)$, where $L(\tau)$ denotes the trajectory length (number of interaction turns).

\subsection{Success Rate under Horizon Constraints}
\label{sec:prelim_metrics}

Let $r(\tau) \in \{0, 1\}$ denote the terminal reward of trajectory $\tau$, where $r(\tau) = 1$ indicates successful task completion. During training, we impose an \emph{interaction budget} (or \emph{horizon constraint}) $K$: a trajectory is forcibly terminated if it reaches $K$ steps without task completion, receiving $r(\tau) = 0$.

For a task distribution $\mathcal{T}$, policy $\pi_\theta$, and horizon constraint $K$, we define the \emph{success rate} as:
\begin{equation}
    \text{SR}(K; \pi_\theta, \mathcal{T}) = \mathbb{E}_{u \sim \mathcal{T}, \tau \sim \pi_\theta(\cdot | u, K)} \left[ r(\tau) \right],
    \label{eq:success_rate}
\end{equation}
where the notation $\tau \sim \pi_\theta(\cdot | u, K)$ indicates that the trajectory is sampled under horizon constraint $K$. The horizon constraint thus defines the maximum number of interaction turns the agent may use per episode, directly affecting both task completion probability and computational cost.

\subsection{Horizon Scheduling in Curriculum Learning}
\label{sec:prelim_horizon}

Existing curriculum-based methods adjust the horizon constraint $K_t$ at training step $t$ according to predetermined schedules:

\textbf{Linear schedule}~\citep{xi2025agentgymrl}: The horizon increases by a fixed amount per training step:
\begin{equation}
    K_t = \min\left(K_{\min} + \delta \cdot t, \, K_{\max}\right),
    \label{eq:linear_schedule}
\end{equation}
where $\delta > 0$ is the increment rate, and $K_{\min}, K_{\max}$ are the lower and upper bounds.

\textbf{Multiplicative schedule}~\citep{shen2025tti}: The horizon is set to a growing integer multiple of $K_{\min}$ at stage boundaries:
\begin{equation}
    K_t = \min\left(K_{\min} \cdot \left(\left\lfloor t / T_{\text{stage}} \right\rfloor + 1\right), \, K_{\max}\right),
    \label{eq:mult_schedule}
\end{equation}
where $T_{\text{stage}}$ is the stage duration. \citet{shen2025tti} write this as $h_i = \mathrm{clip}(h_{\min} \cdot i, h_{\max})$ for iteration $i$, and adopt it over the additive alternative $h_i = \mathrm{clip}(h_{\min} + i, h_{\max})$.

Both schedules are \emph{open-loop}: $K_t$ depends solely on the training step $t$ and is independent of the agent's actual performance or the task distribution's complexity. In contrast, our method adapts $K_t$ based on observed trajectory statistics, forming a \emph{closed-loop} controller.

A complete notation summary is provided in Appendix~\ref{app:notation}.

\section{Method}
\label{sec:method}

\begin{figure*}[t]
    \centering
    \includegraphics[width=0.95\textwidth]{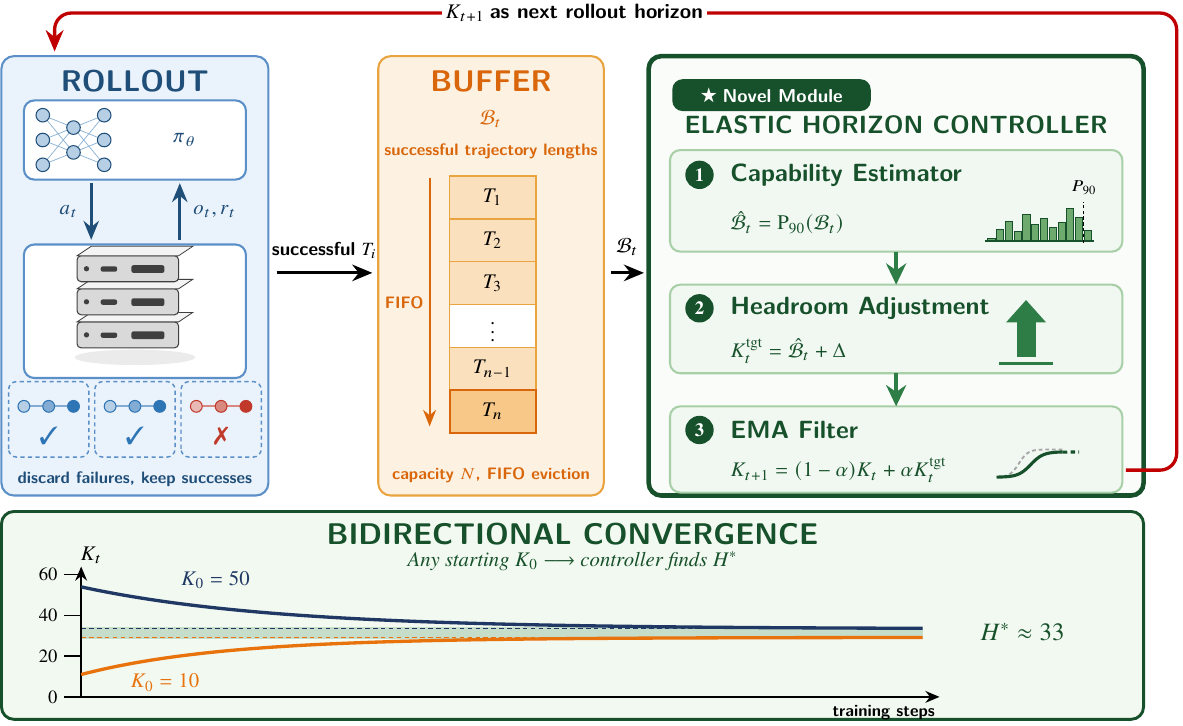}
    \caption{
        \textbf{Elastic Horizon overview.}
        Successful trajectories from \textsc{Rollout} populate the FIFO \textsc{Buffer} $\mathcal{B}_t$ of trajectory lengths.
        The \textsc{Elastic Horizon Controller} (right) maps the buffer to the next horizon in three stages: a percentile \emph{capability estimator} $\hat B_t = \text{P}_{90}(\mathcal{B}_t)$, an additive \emph{headroom adjustment} $K^{\text{tgt}}_t = \hat B_t + \Delta$, and an \emph{EMA filter} $K_{t+1} = (1-\alpha)\,K_t + \alpha\,K^{\text{tgt}}_t$.
        The output $K_{t+1}$ feeds back as the next rollout's horizon (red arrow), forming a closed loop.
        The bottom panel shows the controller's key property: \emph{bidirectional convergence}---from any initialization $K_0$, the horizon settles around the effective frontier $H^*$.
    }
    \label{fig:method}
\end{figure*}

We define the \emph{effective interaction frontier} that motivates our design (Section~\ref{sec:method_hypothesis}), derive principled design requirements for a closed-loop controller (Section~\ref{sec:method_design}), present the capability boundary estimation mechanism (Section~\ref{sec:method_estimation}), and provide the complete Elastic Horizon algorithm (Section~\ref{sec:method_algorithm}).

\subsection{The Effective Interaction Frontier}
\label{sec:method_hypothesis}

Open-loop horizon schedules implicitly assume that performance improves monotonically with interaction budget. The fixed-horizon sweeps in Section~\ref{sec:exp_saturation} (Figure~\ref{fig:sweep}) refute this assumption: success rate plateaus once the interaction budget exceeds a task-dependent threshold, after which additional budget yields no measurable gain in success rate while training compute continues to grow linearly.

We call this threshold the \emph{effective interaction frontier} and denote it $H^*(\mathcal{T}, \theta)$, where $\mathcal{T}$ is the task distribution and $\theta$ the policy parameters. The frontier is the operationally relevant boundary: training with $K > H^*$ pays the linear cost of additional steps without raising success rate, while training with $K < H^*$ caps the agent below its achievable performance. We treat $H^*$ as an empirical quantity defined by the plateau in Figure~\ref{fig:sweep} rather than introducing further formalism, and design our method to track it.

The frontier is \emph{dynamic}: it depends on both $\mathcal{T}$ and the evolving $\theta$. As the agent improves, $H^*$ may shift---a more capable agent can solve harder tasks that require longer interaction sequences. This time-varying property is what makes a fixed $K$ inadequate and motivates a \emph{closed-loop} controller that re-estimates $H^*$ throughout training.

We do not assume that $H^*$ is directly observable. Instead, we use the agent's successful-trajectory length distribution as a behavioral surrogate (Section~\ref{sec:method_estimation}), and validate empirically (Section~\ref{sec:exp_convergence}) that the controller's converged horizon lies within the saturation band identified in Section~\ref{sec:exp_saturation}.

\subsection{Design Requirements for Closed-Loop Control}
\label{sec:method_design}

Given the goal of tracking $H^*$, we identify three design requirements:

\textbf{R1: Capability-driven signal.} The controller should adjust the horizon based on the agent's demonstrated capability, not training progress. Open-loop schedules violate this by conditioning on step $t$ rather than performance.

\textbf{R2: Robust estimation.} The capability signal must be robust to trajectory-level noise (e.g., anomalously long trajectories from inefficient exploration) while remaining sensitive to genuine capability changes.

\textbf{R3: Bidirectional adaptation.} The controller must expand the horizon when capability grows and contract when approaching saturation. Monotonic-only schedules cannot converge to $H^*$ from above.

Elastic Horizon satisfies these requirements through: (R1) using successful trajectory lengths as the capability signal; (R2) employing percentile-based estimation with smoothing; (R3) allowing both expansion and contraction based on the signal.

\subsection{Capability Boundary Estimation}
\label{sec:method_estimation}

The core mechanism of Elastic Horizon is estimating the agent's current capability boundary from trajectory data. We maintain a buffer $\mathcal{B}$ of successful trajectory lengths and compute an upper percentile as the boundary estimate.

\paragraph{Why successful trajectories?}
Successful trajectories directly measure the agent's ability to complete tasks within a given number of steps. The distribution of successful trajectory lengths $F_L$ encodes how much interaction budget the agent \emph{actually needs} to solve tasks it can solve. This provides a grounded capability signal, unlike training loss or reward statistics which conflate capability with task difficulty.

\paragraph{Percentile selection.}
We use the \textbf{90th percentile (P90)} of successful trajectory lengths:
\begin{equation}
    \hat{B}_t = \hat{F}_L^{-1}(0.9) = \text{P90}(\mathcal{B}_t),
    \label{eq:p90}
\end{equation}
where $\hat{F}_L(x)$ is the empirical cumulative distribution function (CDF) of trajectory lengths in buffer $\mathcal{B}_t$.

This choice balances two competing objectives:

\emph{Sensitivity}: The estimate should capture near-maximal capability, providing sufficient horizon for harder task instances. The mean $\mathbb{E}[L]$ underestimates this, as it reflects average behavior rather than capability limits.

\emph{Robustness}: The estimate should filter outliers from inefficient exploration or random walks. The maximum $\max_i L_i$ is sensitive to a single anomalously long trajectory.

P90 achieves both: it captures the capability frontier (90\% of successful trajectories complete within this length) while filtering the top 10\% of potential outliers. From an order statistics perspective, high quantiles provide robust location estimates for heavy-tailed distributions. Agent trajectory lengths are commonly heavy-tailed, as we verify empirically in Appendix~\ref{app:distribution}.

\paragraph{Exploration headroom.}
To encourage attempting tasks slightly beyond current capability, we add a fixed headroom $\Delta$:
\begin{equation}
    K_t^{\text{target}} = \hat{B}_t + \Delta.
    \label{eq:headroom}
\end{equation}
The headroom provides slack for exploring longer trajectories necessary for harder tasks, preventing premature convergence to a suboptimal frontier. This parallels the ``zone of proximal development'' concept in curriculum learning~\citep{bengio2009curriculum}: the agent should be challenged slightly beyond demonstrated competence.

\paragraph{EMA smoothing.}
Raw percentile estimates fluctuate due to sampling variance. We apply exponential moving average (EMA) smoothing to the horizon update:
\begin{equation}
    K_{t+1} = (1-\alpha) \cdot K_t + \alpha \cdot \text{clip}(K_t^{\text{target}}, K_{\min}, K_{\max}),
    \label{eq:ema_update}
\end{equation}
where $\alpha \in (0, 1)$ controls adaptation speed. Small $\alpha$ yields stable but slow adaptation; large $\alpha$ yields responsive but potentially oscillatory behavior. 

\subsection{The Elastic Horizon Algorithm}
\label{sec:method_algorithm}

Algorithm~\ref{alg:elastic_horizon} presents the complete training procedure integrating the above components.

\begin{algorithm}[t]
\caption{Elastic Horizon Training}
\label{alg:elastic_horizon}
\begin{algorithmic}[1]
\Require Policy $\pi_{\theta_0}$, task distribution $\mathcal{T}$
\Require Initial horizon $K_0$, bounds $[K_{\min}, K_{\max}]$
\Require EMA coefficient $\alpha$, headroom $\Delta$, buffer size $N$
\State Initialize buffer $\mathcal{B} \leftarrow \emptyset$, horizon $K \leftarrow K_0$
\For{training step $t = 1, 2, \ldots, T$}
    \State \textcolor{gray}{\textit{// Trajectory collection under current horizon}}
    \State Sample trajectories $\{\tau_i\}_{i=1}^{M}$ with budget $K$
    \State \textcolor{gray}{\textit{// Policy update}}
    \State $\theta_t \leftarrow \textsc{RLUpdate}(\theta_{t-1}, \{\tau_i\})$
    \State \textcolor{gray}{\textit{// Buffer update (FIFO)}}
    \For{each $\tau_i$ with $r(\tau_i) = 1$}
        \State $\mathcal{B}.\textsc{Append}(L(\tau_i))$
        \If{$|\mathcal{B}| > N$} $\mathcal{B}.\textsc{PopFirst}()$
        \EndIf
    \EndFor
    \State \textcolor{gray}{\textit{// Closed-loop horizon update}}
    \If{$|\mathcal{B}| \geq N_{\min}$}
        \State $\hat{B} \leftarrow \text{P90}(\mathcal{B})$ \Comment{Capability estimate}
        \State $K^{\text{raw}} \leftarrow \text{clip}(\hat{B} + \Delta, K_{\min}, K_{\max})$
        \State $K \leftarrow (1 - \alpha) \cdot K + \alpha \cdot K^{\text{raw}}$ \Comment{EMA}
    \EndIf
\EndFor
\end{algorithmic}
\end{algorithm}

\paragraph{Cold start handling.}
When successful samples are scarce ($|\mathcal{B}| < N_{\min}$), the P90 estimate is unreliable. The controller maintains the current horizon until sufficient data accumulates. This conservative strategy prevents erratic behavior during early training when exploration dominates.

\paragraph{Computational overhead.}
Elastic Horizon adds negligible cost: maintaining a buffer of integers, computing a percentile, and performing EMA update require $O(N)$ operations per step, which is insignificant compared to trajectory sampling and gradient computation.

\paragraph{Comparison with open-loop schedules.}
Table~\ref{tab:method_comparison} summarizes key differences. Open-loop methods depend solely on training step $t$ and increase monotonically to $K_{\max}$. Elastic Horizon depends on agent performance and adapts bidirectionally, automatically converging toward $H^*$.

\begin{table}[t]
\centering
\caption{Comparison of horizon scheduling approaches.}
\label{tab:method_comparison}
\footnotesize
\setlength{\tabcolsep}{3pt}
\begin{tabular}{lccc}
\toprule
& \textbf{ScalingInter} & \textbf{TTI} & \textbf{Elastic (Ours)} \\
\midrule
Control & Open-loop & Open-loop & Closed-loop \\
Signal & Step $t$ & Step $t$ & P90 of $L_{\text{succ}}$ \\
Direction & $\uparrow$ only & $\uparrow$ only & Bidirectional \\
Converges to & $K_{\max}$ & $K_{\max}$ & $\approx H^*$ \\
Hparams & $\delta$, stages & $K_{\min}$, $T_{\text{stg}}$ & $\alpha$, $\Delta$ \\
\bottomrule
\end{tabular}
\end{table}



\section{Experiments}
\label{sec:experiments}

We design experiments to answer three research questions:
\textbf{RQ1}: Does interaction horizon saturation exist?
\textbf{RQ2}: Can Elastic Horizon automatically discover the effective frontier?
\textbf{RQ3}: How does Elastic Horizon compare to open-loop baselines?

\subsection{Experimental Setup}
\label{sec:exp_setup}

\paragraph{Environments.}
We evaluate on two challenging multi-turn agent benchmarks:

\textbf{AppWorld}~\citep{trivedi2024appworld} is a high-fidelity execution environment simulating 9 daily applications (Amazon, Venmo, Spotify, Gmail, etc.) with 457 APIs. Tasks require agents to coordinate across multiple apps through API calls. We report task goal completion rate.

\textbf{BFCL v3}~\citep{patil2025bfcl} (Berkeley Function Calling Leaderboard) evaluates multi-turn function calling across 8 API domains. The multi-turn split contains 1,000 test cases requiring stateful interactions. We report overall success rate.

Both benchmarks use binary sparse rewards and enforce a maximum trajectory length of 50 steps.

\paragraph{Baselines.}
We compare against:
\textbf{GRPO (Fixed $K$)}: Group Relative Policy Optimization~\citep{shao2024deepseekmath} with static horizon constraints ($K=15$ and $K=50$).
\textbf{TTI}~\citep{shen2025tti}: Multiplicative horizon schedule ($15 \to 20 \to 30 \to 50$).
\textbf{ScalingInter}~\citep{xi2025agentgymrl}: Linear horizon schedule from $K_{\min}=10$ to $K_{\max}=50$.

\paragraph{Training details.}
We use GRPO with KL coefficient zero, training for 200 steps with batch size 16 and learning rate $1 \times 10^{-6}$ on both Qwen2.5-7B and Qwen2.5-14B. Our training pipeline is built on the AgentEvolver agent training framework~\citep{zhai2025agentevolver}. For Elastic Horizon: $K_0 = 15$, $\alpha = 0.1$, $\Delta = 10$, buffer size $N = 100$. We report mean@8 (average over 8 rollouts). Full details in Appendix~\ref{app:exp_details}.

\subsection{RQ1: Does Horizon Saturation Exist?}
\label{sec:exp_saturation}

To verify interaction horizon saturation, we train separate models with fixed horizon constraints, sweeping $K$ from $10$ up to $80$ on both benchmarks.

\begin{figure*}[t]
    \centering
    \includegraphics[width=0.95\textwidth]{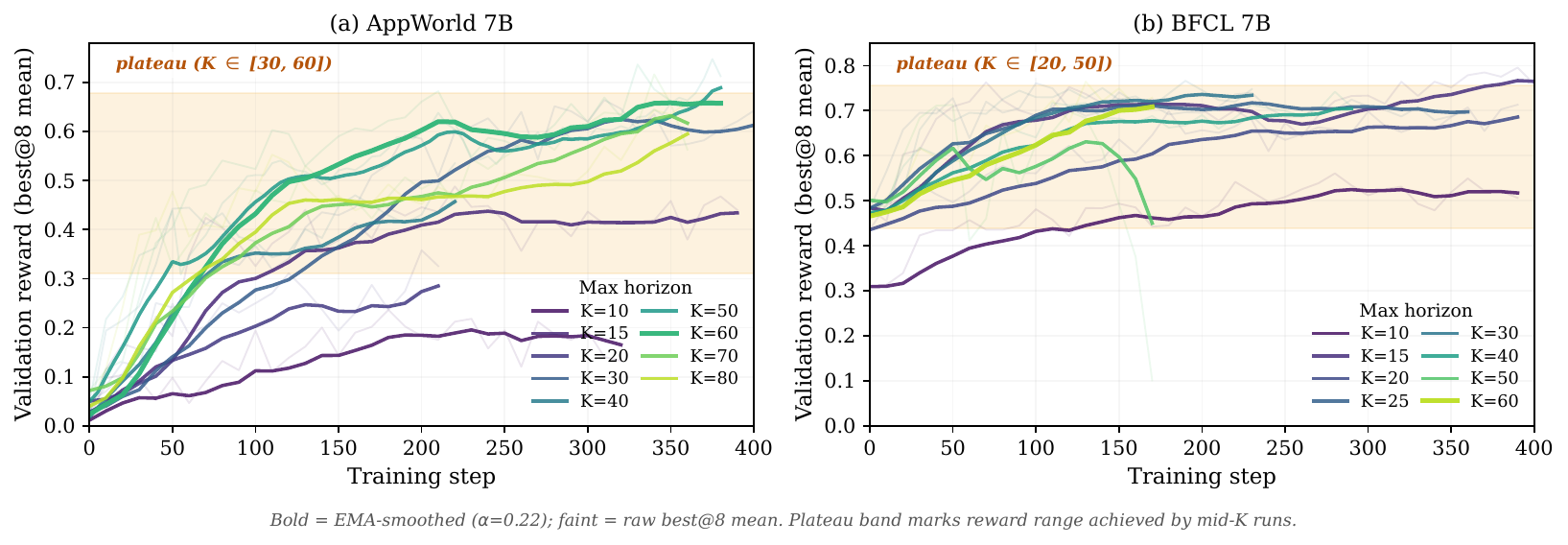}
    \caption{
        \textbf{Training dynamics under fixed horizons.}
        Validation reward (best@8 mean) versus training step for fixed-horizon GRPO runs on AppWorld 7B (a) and BFCL 7B (b).
        Curves are colored by $K$ on a viridis scale; bold lines are EMA-smoothed ($\alpha{=}0.22$), faint lines are raw measurements.
        On both benchmarks, performance saturates within a benchmark-specific plateau band (orange shading): $K \in [30, 60]$ on AppWorld and $K \in [20, 50]$ on BFCL.
        Beyond the plateau, additional horizon budget produces no measurable gain in reward despite linearly higher per-step cost.
    }
    \label{fig:sweep}
\end{figure*}

Figure~\ref{fig:sweep} reveals a consistent saturation pattern on both benchmarks. In the \emph{under-capacity regime} ($K < H^*$), reward increases substantially with horizon length: AppWorld 7B climbs from $\sim$17\% at $K{=}10$ to $\sim$60\% at $K{=}30$. In the \emph{saturation regime} ($K \approx H^*$), reward plateaus---runs with $K \in [30, 60]$ converge to the same band on AppWorld, and $K \in [20, 50]$ converge on BFCL---despite a 2$\times$ range of per-step cost. In the \emph{noise-dominated regime} ($K > H^*$), additional budget yields no systematic improvement.

The saturation threshold differs between benchmarks: AppWorld, with complex multi-app coordination, exhibits a higher $H^*$ than BFCL's function-calling tasks. This validates that $H^*$ is task-dependent and underscores the need for adaptive horizon control.

\subsection{RQ2: Automatic Frontier Discovery}
\label{sec:exp_convergence}

We test Elastic Horizon's ability to adapt the horizon based on agent capability by running from two extreme initializations: $K_0 = 10$ (under-capacity) and $K_0 = 50$ (over-capacity). Figure~\ref{fig:sub2} (right panel of Figure~\ref{fig:total_figure}) shows the horizon trajectories. Open-loop baselines (ScalingInter, TTI) increase monotonically to $K_{\max} = 50$ regardless of agent behavior. Elastic Horizon behaves differently: from $K_0 = 10$, the horizon expands as the buffer fills with longer successful trajectories; from $K_0 = 50$, it contracts once the P90 estimate stabilizes. In both cases, the horizon stabilizes inside the saturation band identified in RQ1 (Figure~\ref{fig:sweep}). The exact stable value differs between runs (since it depends on hyperparameters $\alpha$ and $\Delta$), but in neither case does the horizon reach $K_{\max}$.

We interpret this as evidence that Elastic Horizon adapts the horizon based on the agent's demonstrated trajectory length distribution rather than training step, which is the property that distinguishes it from open-loop schedules.

\subsection{RQ3: Comparison with Baselines}
\label{sec:exp_main}

Table~\ref{tab:main_results} summarizes the main results.

\begin{table}[t]
\centering
\caption{
    Performance comparison. Upper: zero-shot backbone models. Lower: training with Qwen2.5-7B and Qwen2.5-14B.
    $^*$: training instability. Best in \textbf{bold}.
}
\label{tab:main_results}
\footnotesize
\setlength{\tabcolsep}{2pt}
\begin{tabular}{lcccc}
\toprule
\textbf{Method} & \multicolumn{2}{c}{\textbf{AppWorld (\%)}} & \multicolumn{2}{c}{\textbf{BFCL (\%)}} \\
\midrule
\multicolumn{5}{l}{\textit{Zero-shot Baselines}} \\
Qwen2.5-7B      & \multicolumn{2}{c}{2.08}  & \multicolumn{2}{c}{30.12} \\
Qwen2.5-14B     & \multicolumn{2}{c}{18.00} & \multicolumn{2}{c}{41.62} \\
Qwen2.5-32B     & \multicolumn{2}{c}{39.06} & \multicolumn{2}{c}{49.63} \\
Qwen3-235B-A22B & \multicolumn{2}{c}{33.85} & \multicolumn{2}{c}{60.88} \\
\midrule
\textit{Training (7B)} & Mean@8 & Best@8 & Mean@8 & Best@8 \\
\cmidrule(r){1-1} \cmidrule(lr){2-3} \cmidrule(l){4-5}
GRPO ($K=15$)        & 21.49 & 42.40 & 66.87 & 71.62 \\
GRPO ($K=50$)        & 31.14 & 49.20 & 60.37$^*$ & 68.32$^*$ \\
TTI                  & 37.28 & 46.14 & 55.75 & 71.22 \\
ScalingInter         & 36.18 & 55.10 & 69.12 & 73.93 \\
\textbf{Elastic Horizon} & \textbf{39.47} & \textbf{56.87} & \textbf{71.62} & \textbf{80.45} \\
\midrule
\textit{Training (14B)} & Mean@8 & Best@8 & Mean@8 & Best@8 \\
\cmidrule(r){1-1} \cmidrule(lr){2-3} \cmidrule(l){4-5}
GRPO ($K=15$)        & 56.57 & 71.16 & 68.80 & 74.13 \\
GRPO ($K=50$)        & 65.78 & 80.63 & 64.12 & 73.61 \\
TTI                  & 60.08 & 77.73 & 70.54 & 76.35 \\
ScalingInter         & 67.54 & 82.54 & 70.17 & 78.85 \\
\textbf{Elastic Horizon} & \textbf{77.85} & \textbf{90.42} & \textbf{73.62} & \textbf{80.49} \\
\bottomrule
\end{tabular}
\end{table}

\paragraph{Static horizon dilemma.}
GRPO ($K{=}15$) vs.\ GRPO ($K{=}50$) illustrates the challenge of fixed-horizon training. On AppWorld, longer horizons help (complex multi-app tasks benefit from extended interaction); on BFCL they hurt, with training instability (marked $^*$), confirming that exceeding $H^*$ degrades learning efficiency.

\paragraph{Open-loop limitations.}
TTI and ScalingInter give inconsistent results across benchmarks: both reach $K_{\max}$ regardless of task complexity, leaving them unable to adapt when the optimal horizon differs across domains.

\paragraph{Elastic Horizon robustness.}
Elastic Horizon achieves the best results on both benchmarks at both 7B and 14B (e.g., AppWorld Best@8${=}90.42\%$ at 14B). The trained 7B model surpasses zero-shot Qwen2.5-32B (39.06\%) and Qwen3-235B-A22B (33.85\%) on AppWorld.

\begin{figure}[t]
    \centering
    \includegraphics[width=\linewidth]{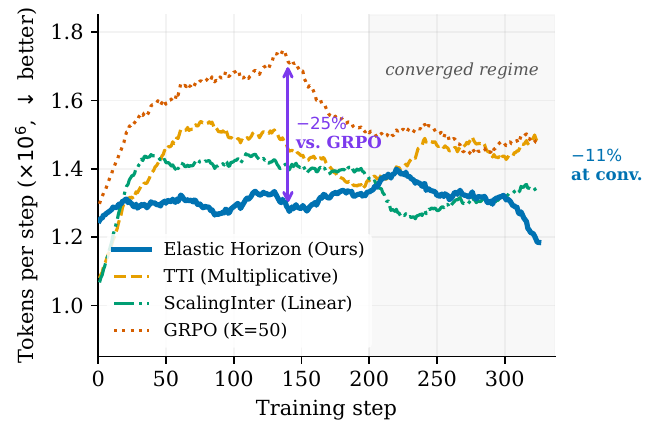}
    \caption{
        \textbf{Per-step token consumption on AppWorld (14B).}
        Elastic Horizon consumes $-$25\% vs.\ GRPO at the mid-training peak (step ${\sim}140$) and $-$11\% vs.\ the worst baseline at convergence (step ${\geq}200$). Cumulative-token and BFCL counterparts in Appendix~\ref{app:efficiency}.
    }
    \label{fig:efficiency}
\end{figure}

\paragraph{Sample efficiency.}
Elastic Horizon saves tokens at two regimes (Figure~\ref{fig:efficiency}): at mid-training (steps 100--170) when GRPO peaks, EH has already converged and saves 25\% per-step tokens; at convergence (step ${\geq}200$), EH still uses 11\% fewer per-step tokens than the worst baseline. These add up to a 28\% cumulative-token saving at SR${=}50\%$ on AppWorld 7B (Appendix~\ref{app:efficiency}). Under tight compute budgets, the mid-training saving is the more consequential.

\section{Conclusion}
\label{sec:conclusion}

We introduced \emph{Elastic Horizon}, a closed-loop controller that sets the interaction budget of an agentic RL run from the agent's own demonstrated behavior, estimating the effective interaction frontier $H^*$ from the 90th percentile of successful trajectory lengths. Three findings emerge. First, fixed-horizon training saturates beyond a task-dependent $H^*$: on AppWorld and BFCL, a twofold range of interaction budgets converges to the same reward band while per-step cost keeps growing. Second, because the controller reads capability rather than training step, it both expands and contracts the horizon, and settles inside the saturation band from under- and over-capacity initializations alike, behavior that monotonically increasing schedules cannot produce. Third, this costs nothing in quality: Elastic Horizon attains the best success rates across 7B and 14B backbones on both benchmarks while cutting per-step trajectory tokens by up to 25\%.

The broader point is that the interaction budget need not be a hyperparameter. Open-loop curricula ask practitioners to commit in advance to a growth rate and a maximum horizon, quantities that depend on the agent and the environment and are therefore unknown before training begins. Elastic Horizon replaces that commitment with a measurement, so that how far to scale the horizon---and where to stop---is determined by the training run itself. We expect the same treatment, allocating from demonstrated capability in both directions, to be applicable to other budgets in agent training beyond the interaction horizon.



\section*{Limitations}
\label{sec:limitations}

We discuss several limitations of our work that suggest directions for future research.

\paragraph{Cold start with sparse success.}
Elastic Horizon relies on successful trajectory statistics for capability estimation. In early training or on extremely difficult tasks, successful trajectories may be rare, leading to unreliable P90 estimates. Our current mitigation---maintaining the initial horizon until a minimum buffer is reached---is conservative but may delay adaptation. Future work could explore alternative signals such as partial progress indicators or trajectory length distributions from all rollouts (not just successful ones).

\paragraph{Single global horizon.}
We adjust a single global horizon $K_t$ applied uniformly across all task instances. For heterogeneous task distributions with wide difficulty variance, this may be suboptimal: easy tasks receive unnecessarily long horizons while hard tasks may still be under-budgeted. Instance-level or difficulty-conditioned horizon allocation could improve efficiency further, though at the cost of additional complexity.

\paragraph{Heuristic nature of P90.}
The choice of the 90th percentile is empirically motivated and supported by ablation studies, but lacks a first-principles derivation. While we provide intuitions from order statistics and curriculum learning theory, a more rigorous theoretical foundation for optimal percentile selection remains an open question. The ablation results suggest that performance is relatively robust across the P75--P95 range, but the optimal choice may be task-dependent.

\paragraph{Proxy efficiency metric.}
We use cumulative trajectory tokens as a proxy for training compute. While more informative than raw interaction counts, this metric does not capture all cost factors such as environment latency, parallel sampling efficiency, or accelerator utilization. Actual wall-clock time savings may differ from token-based estimates.

\section*{Ethical Considerations}

Our proposed method, Elastic Horizon, explicitly addresses the computational overhead associated with long-horizon agentic tasks. By automatically detecting the effective interaction frontier, our approach reduces per-step trajectory tokens by up to 25\% during the most expensive phase of training and 11\% at convergence on AppWorld, compared to fixed-horizon baselines. This contributes to the goals of ``Green AI'' by lowering the energy consumption and carbon footprint required to train capable agents. By demonstrating that smaller models (e.g., 7B parameters) can achieve high performance through adaptive horizon control, our work also helps make agent research more accessible to institutions with limited computational resources.

While we focus on training efficiency, we acknowledge that more capable LLM agents carry inherent risks of misuse. Our method does not introduce specific new capabilities that would uniquely exacerbate these risks beyond the general progress of the field; it is a training-efficiency technique applicable to any agentic RL setup. We use only publicly released benchmarks (AppWorld, BFCL) and base models (Qwen2.5), and our experiments do not involve human subjects or sensitive data.



\bibliography{references}


\appendix

\section{Notation Summary}
\label{app:notation}

Table~\ref{tab:notation_full} provides a complete summary of notation used throughout the paper.

\begin{table}[t]
\centering
\caption{Complete notation summary.}
\label{tab:notation_full}
\footnotesize
\setlength{\tabcolsep}{2pt}
\resizebox{\columnwidth}{!}{%
\begin{tabular}{@{}clc@{}}
\toprule
\textbf{Symbol} & \textbf{Description} & \textbf{Defined in} \\
\midrule
\multicolumn{3}{l}{\textit{POMDP Formulation}} \\
$\mathcal{M}$ & POMDP tuple & Sec.~\ref{sec:prelim_pomdp} \\
$\mathcal{U}$ & Instruction space & Sec.~\ref{sec:prelim_pomdp} \\
$\mathcal{S}$ & State space & Sec.~\ref{sec:prelim_pomdp} \\
$\mathcal{A}$ & Action space & Sec.~\ref{sec:prelim_pomdp} \\
$\mathcal{O}$ & Observation space & Sec.~\ref{sec:prelim_pomdp} \\
$T$ & Transition function & Sec.~\ref{sec:prelim_pomdp} \\
$R$ & Reward function & Sec.~\ref{sec:prelim_pomdp} \\
$\pi_\theta$ & Policy w/ params $\theta$ & Sec.~\ref{sec:prelim_pomdp} \\
$\mathcal{T}$ & Task distribution & Sec.~\ref{sec:prelim_metrics} \\
\midrule
\multicolumn{3}{l}{\textit{Trajectory and Metrics}} \\
$\tau$ & Trajectory & Sec.~\ref{sec:prelim_pomdp} \\
$L(\tau)$ & Trajectory length (turns) & Sec.~\ref{sec:prelim_pomdp} \\
$r(\tau)$ & Terminal reward (1=succ, 0=fail) & Sec.~\ref{sec:prelim_metrics} \\
$\text{SR}(K; \pi_\theta, \mathcal{T})$ & Success rate under horizon $K$ & Sec.~\ref{sec:prelim_metrics} \\
\midrule
\multicolumn{3}{l}{\textit{Horizon Control}} \\
$K_t$ & Horizon at step $t$ & Sec.~\ref{sec:prelim_horizon} \\
$K_{\min}$ & Minimum horizon bound & Sec.~\ref{sec:prelim_horizon} \\
$K_{\max}$ & Maximum horizon bound & Sec.~\ref{sec:prelim_horizon} \\
$H^*$ & Effective frontier & Sec.~\ref{sec:method_hypothesis} \\
\midrule
\multicolumn{3}{l}{\textit{Elastic Horizon}} \\
$\mathcal{B}$ & Success buffer (trajectory lengths) & Sec.~\ref{sec:method_algorithm} \\
$\hat{B}_t$ & Capability boundary estimate & Sec.~\ref{sec:method_estimation} \\
$\text{P90}(\mathcal{B})$ & 90th percentile of $\mathcal{B}$ & Sec.~\ref{sec:method_estimation} \\
$F_L$ & CDF of success lengths & Sec.~\ref{sec:method_estimation} \\
$\alpha$ & EMA smoothing coefficient & Sec.~\ref{sec:method_algorithm} \\
$\Delta$ & Headroom & Sec.~\ref{sec:method_estimation} \\
$N$ & Buffer capacity & App.~\ref{app:exp_details} \\
$N_{\min}$ & Min.\ buffer for activation & App.~\ref{app:exp_details} \\
\bottomrule
\end{tabular}%
}
\end{table}

\section{Theoretical Analysis}
\label{app:theory}

This appendix collects formal discussion of the design choices in Elastic Horizon. We do not claim a full theoretical analysis: the controller is fundamentally a closed-loop heuristic whose primary justification is empirical (Section~\ref{sec:experiments}). What we make explicit here are (i) the statistical motivation for using the 90th percentile of successful trajectory lengths as a capability boundary estimator (Section~\ref{app:p90_theory}); and (ii) the stability behavior we rely on for the EMA update, together with the limitations of treating it as a convergence result (Section~\ref{app:convergence}).

\subsection{Statistical Motivation for the P90 Estimator}
\label{app:p90_theory}

The capability boundary estimator $\hat{B}_t = \text{P90}(\mathcal{B}_t)$ is motivated by two desiderata: it should reflect near-maximal demonstrated capability while remaining robust to occasional very long successful trajectories produced by inefficient exploration. Both desiderata follow from standard order-statistics intuition; we do not derive a closed-form relationship between $\hat{B}_t$ and the frontier $H^*$.

\paragraph{Heavy-tail robustness.}
Successful trajectory lengths are right-skewed in our experiments (Appendix~\ref{app:distribution}, Table~\ref{tab:quantiles}). Under heavy tails, the sample maximum has variance
\begin{equation}
    \text{Var}[\max_i L_i] = O(n^{2/\alpha - 2}),
\end{equation}
for Pareto tails with index $\alpha$, which diverges for $\alpha < 2$. By contrast, sample quantiles in the body of the distribution---including the 90th percentile under standard regularity conditions---have variance $O(1/n)$. The P90 therefore provides a far more stable boundary estimate than the maximum when occasional very long successful trajectories occur.

\paragraph{Choice of P90 vs.\ other quantiles.}
The choice of the 90th percentile, rather than P75 or P95, is empirically motivated. Lower quantiles (mean, P50) underestimate near-maximal capability and lead to overly conservative horizons that truncate harder task instances; higher quantiles approach the maximum and lose the robustness benefit above. The ablation in Appendix~\ref{app:ablation} confirms that performance is relatively insensitive across the P75--P95 range, consistent with the order-statistics intuition that any high quantile gives a stable, near-maximal estimate. We do not claim that P90 is optimal in any formal sense.

\paragraph{Connection to the frontier.}
We do not establish a closed-form bound between $\hat{B}_t$ and $H^*$. The intended interpretation is qualitative: $\hat{B}_t$ captures the horizon at which $90\%$ of the agent's currently solvable tasks complete, so $\hat{B}_t + \Delta$ is a budget large enough to admit nearly all current successes plus some slack for exploration of harder instances. Whether this surrogate tracks the true frontier is an empirical question; the convergence experiments in Section~\ref{sec:exp_convergence}, where the controller stabilizes near the same plateau identified by the fixed-horizon sweeps in Section~\ref{sec:exp_saturation}, are the main evidence we offer.

\subsection{Stability of the EMA Update}
\label{app:convergence}

The horizon update rule
\begin{align}
    K_{t+1} &= (1-\alpha) K_t + \alpha K^{\text{raw}}_t,\\
    K^{\text{raw}}_t &= \text{clip}(\hat{B}_t + \Delta, K_{\min}, K_{\max}),
\end{align}
is a standard exponential moving average over the clipped target $K^{\text{raw}}_t$. Its stability is a property of the EMA filter rather than a novel result, but it is worth stating the form we rely on, together with what it does \emph{not} say.

\paragraph{Geometric attraction toward a stationary target.}
Suppose the input sequence $\{K^{\text{raw}}_t\}$ converges to a fixed target $K^\dagger$. Subtracting $K^\dagger$ from both sides of the EMA update gives $|K_{t+1} - K^\dagger| \le (1-\alpha)|K_t - K^\dagger| + \alpha |K^{\text{raw}}_t - K^\dagger|$. Iterating yields the standard bound
\begin{multline}
    |K_t - K^\dagger| \le (1-\alpha)^t |K_0 - K^\dagger| \\
    + \alpha \sum_{s=0}^{t-1} (1-\alpha)^{t-1-s} |K^{\text{raw}}_s - K^\dagger|.
\end{multline}
The first term decays geometrically; the second is controlled by the residual estimation error of $K^{\text{raw}}_s$.

\paragraph{Implications for the controller.}
This standard property has two consequences for Elastic Horizon. First, the steady-state horizon depends on the limiting target distribution of $K^{\text{raw}}_t$ (which in turn depends on hyperparameters $\alpha, \Delta$), not on the initial horizon $K_0$. This is consistent with the convergence experiments in Section~\ref{sec:exp_convergence}, where runs starting from $K_0=10$ and $K_0=50$ both stabilize within the saturation band identified in Section~\ref{sec:exp_saturation}. Second, the smoothing constant $\alpha$ trades responsiveness against variance: small $\alpha$ damps sampling noise in $\hat{B}_t$ but tracks capability changes more slowly, as confirmed empirically in Appendix~\ref{app:ablation}.

\paragraph{Limitation: non-stationary targets from policy updates.}
The above analysis assumes the input $K^{\text{raw}}_t$ converges to a fixed target. In Elastic Horizon training, both the policy $\pi_\theta$ and the trajectory-length distribution $F_L$ evolve over time, and the frontier $H^*$ itself is dynamic (Section~\ref{sec:method_hypothesis}). The buffer $\mathcal{B}_t$ uses FIFO eviction precisely so that $\hat{B}_t$ tracks current rather than historical capability; as a consequence, $K^{\text{raw}}_t$ is not a stationary input. The EMA bound above should therefore be read as a description of the controller's \emph{tracking} behavior under slow drift, not a claim of asymptotic convergence to a fixed point. Formal analysis of the joint policy--controller dynamics, including the cold-start and FIFO buffer effects, is left to future work.

\section{Algorithm Details}
\label{app:algorithm}

Algorithm~\ref{alg:elastic_horizon_full} presents the complete Elastic Horizon training procedure with all implementation details.

\begin{algorithm}[h]
\caption{Elastic Horizon Training (Complete)}
\label{alg:elastic_horizon_full}
\begin{algorithmic}[1]
\Require Initial policy $\pi_{\theta_0}$, task distribution $\mathcal{T}$
\Require Initial horizon $K_0$, horizon bounds $K_{\min}, K_{\max}$
\Require EMA coefficient $\alpha$, headroom $\Delta$
\Require Buffer capacity $N$, minimum buffer size $N_{\min}$
\Require Number of training steps $T$, batch size $M$
\State Initialize success buffer $\mathcal{B} \leftarrow \emptyset$
\State Initialize horizon $K \leftarrow K_0$
\For{training step $t = 1, 2, \ldots, T$}
    \State \textcolor{gray}{\textit{// Phase 1: Trajectory Collection}}
    \State Sample task batch $\{u_i\}_{i=1}^{M}$ from $\mathcal{T}$
    \For{$i = 1, \ldots, M$}
        \State Initialize trajectory $\tau_i \leftarrow (u_i)$
        \For{interaction step $k = 1, \ldots, K$}
            \State Sample action $a_k \sim \pi_{\theta_{t-1}}(\cdot | \tau_i)$
            \State Execute action, receive observation $o_k$ and done flag $d_k$
            \State Append to trajectory: $\tau_i \leftarrow \tau_i \cup (a_k, o_k)$
            \If{$d_k = \textsc{True}$}
                \State \textbf{break}
            \EndIf
        \EndFor
        \State Compute reward $r_i \leftarrow r(\tau_i)$ and length $L_i \leftarrow L(\tau_i)$
    \EndFor
    \State \textcolor{gray}{\textit{// Phase 2: Policy Update}}
    \State $\theta_t \leftarrow \textsc{GRPO}(\theta_{t-1}, \{(\tau_i, r_i)\}_{i=1}^{M})$
    \State \textcolor{gray}{\textit{// Phase 3: Buffer Update (FIFO)}}
    \For{$i$ such that $r_i = 1$}
        \State $\mathcal{B}.\textsc{Append}(L_i)$
        \If{$|\mathcal{B}| > N$}
            \State $\mathcal{B}.\textsc{PopFirst}()$
        \EndIf
    \EndFor
    \State \textcolor{gray}{\textit{// Phase 4: Horizon Update (Closed-Loop Control)}}
    \If{$|\mathcal{B}| \geq N_{\min}$}
        \State $\hat{B} \leftarrow \textsc{Percentile}(\mathcal{B}, 90)$ \Comment{Capability boundary estimate}
        \State $K^{\text{raw}} \leftarrow \textsc{Clip}(\hat{B} + \Delta, K_{\min}, K_{\max})$
        \State $K \leftarrow \lfloor (1 - \alpha) \cdot K + \alpha \cdot K^{\text{raw}} \rfloor$ \Comment{EMA smoothing}
    \EndIf
    \State Log $(t, K, \hat{B}, \text{SR}_t)$
\EndFor
\State \Return Trained policy $\pi_{\theta_T}$
\end{algorithmic}
\end{algorithm}

\paragraph{Implementation notes.}
\begin{itemize}[leftmargin=*, itemsep=2pt]
    \item Training is implemented on top of the AgentEvolver agent training framework~\citep{zhai2025agentevolver}. Elastic Horizon is added as a horizon-control module around the rollout loop (Phases 1, 3, and 4 above); the policy update path is left unmodified.
    \item The GRPO update follows~\citet{shao2024deepseekmath} with KL coefficient set to zero.
    \item The buffer $\mathcal{B}$ uses first-in-first-out (FIFO) eviction to maintain a sliding window of the most recent successful trajectory lengths, ensuring the estimate reflects current agent capability rather than historical performance.
    \item The floor operation $\lfloor \cdot \rfloor$ ensures integer horizon values.
    \item During cold start ($|\mathcal{B}| < N_{\min}$), the horizon remains unchanged to avoid unreliable estimates from insufficient samples.
    \item We use $N_{\min} = 20$ to ensure the P90 estimate is based on at least 20 successful trajectories.
\end{itemize}

\section{Extended Experimental Details}
\label{app:exp_details}

\subsection{Hyperparameter Settings}

Table~\ref{tab:hyperparams} provides the complete hyperparameter configuration for all experiments.

\begin{table}[t]
\centering
\caption{Complete hyperparameter settings.}
\label{tab:hyperparams}
\footnotesize
\setlength{\tabcolsep}{2pt}
\begin{tabular}{@{}llc@{}}
\toprule
\textbf{Category} & \textbf{Parameter} & \textbf{Value} \\
\midrule
\multirow{6}{*}{Training} 
    & Optimizer & AdamW \\
    & Learning rate & $1 \times 10^{-6}$ \\
    & Batch size & 16 \\
    & Micro-batch size & 8 \\
    & Grad accum steps & 2 \\
    & Total training steps & 200 \\
\midrule
\multirow{4}{*}{GRPO} 
    & KL coefficient & 0.0 \\
    & Discount factor $\gamma$ & 1.0 \\
    & GAE $\lambda$ & 1.0 \\
    & Clip ratio & 0.2 \\
\midrule
\multirow{7}{*}{Elastic Horizon} 
    & Initial horizon $K_0$ & 15 \\
    & Min.\ horizon $K_{\min}$ & 5 \\
    & Max.\ horizon $K_{\max}$ & 50 \\
    & EMA coefficient $\alpha$ & 0.1 \\
    & Headroom $\Delta$ & 10 \\
    & Buffer size $N$ & 100 \\
    & Min.\ buffer $N_{\min}$ & 20 \\
\midrule
\multirow{3}{*}{Baselines} 
    & Linear inc.\ $\delta$ & 0.2 per step \\
    & TTI growth stages & $15 \to 20 \to 30 \to 50$ \\
    & TTI stage duration & 50 steps \\
\midrule
\multirow{2}{*}{Evaluation} 
    & Rollouts per task & 8 \\
    & Primary metric & mean@8 \\
\bottomrule
\end{tabular}
\end{table}

\subsection{Environment Configuration}

\textbf{AppWorld}~\citep{trivedi2024appworld}: A high-fidelity execution environment simulating 9 daily applications (Amazon, Venmo, Spotify, Gmail, etc.) with 457 APIs. Tasks require agents to coordinate across multiple apps through API calls. We use the standard test split with binary sparse rewards (1 for task completion, 0 otherwise).

\textbf{BFCL v3}~\citep{patil2025bfcl}: The Berkeley Function Calling Leaderboard evaluates multi-turn function calling across 8 API domains. The multi-turn split contains 1,000 test cases requiring stateful interactions with multiple function calls per task.

Both benchmarks enforce a maximum trajectory length of 50 steps. Trajectories exceeding the horizon constraint $K_t$ are terminated with reward 0.

\subsection{Baseline Implementations}

\textbf{Linear schedule} (ScalingInter-RL style):
\begin{equation}
    K_t = \min(K_{\min} + \delta \cdot t, K_{\max}), \quad \text{with } \delta = 0.2.
\end{equation}
This reaches $K_{\max} = 50$ at step $t = 200$.

\textbf{Multiplicative schedule} (TTI style): The horizon follows $K \in \{15, 20, 30, 50\}$ at training steps $\{0, 50, 100, 150\}$. This follows the stage-wise multiplicative curriculum of~\citet{shen2025tti}, who raise the horizon at successive stage boundaries rather than by a fixed per-step increment; we start from the same initial horizon as the conservative fixed-horizon baseline ($K{=}15$) and cap the schedule at $K_{\max} = 50$.

\textbf{Fixed horizon baselines}: GRPO with $K = 15$ (conservative) and $K = 50$ (maximum), representing the extremes of fixed-horizon training.

\subsection{Computational Resources}

All experiments were conducted on a single compute node with 8 AI accelerator devices per training run. A 7B-parameter Elastic Horizon training run (200 training steps) requires approximately 320 device-hours of wall-clock time; a 14B run requires approximately 680 device-hours under the same settings. The Elastic Horizon controller itself adds negligible overhead ($<0.1\%$) compared to trajectory sampling and gradient computation.

\subsection{Statistical Significance}
\label{app:multiseed}

All main experiments are repeated with 3 random seeds. Table~\ref{tab:main_results_std} reports mean@8 $\pm$ standard deviation across the 3 seeds for the 7B configuration, complementing the point estimates in Table~\ref{tab:main_results}. We do not report multi-seed variance for 14B due to compute constraints: each 14B training run consumes approximately 680 device-hours on 8 accelerator devices ($\sim$85 hours of wall-clock time per run), making 3-seed replication prohibitive.

\begin{table}[t]
\centering
\caption{
    Mean@8 with standard deviations across 3 random seeds (Qwen2.5-7B). Best in \textbf{bold}. $^*$: training instability (highest variance, $\pm 4.52$).
}
\label{tab:main_results_std}
\small
\setlength{\tabcolsep}{4pt}
\begin{tabular}{lcc}
\toprule
\textbf{Method} & \textbf{AppWorld} & \textbf{BFCL} \\
\midrule
GRPO ($K{=}15$)        & 21.49 $\pm$ 2.14 & 66.87 $\pm$ 1.83 \\
GRPO ($K{=}50$)        & 31.14 $\pm$ 3.07 & 60.37 $\pm$ 4.52$^*$ \\
TTI                    & 37.28 $\pm$ 2.91 & 55.75 $\pm$ 3.68 \\
ScalingInter           & 36.18 $\pm$ 2.53 & 69.12 $\pm$ 2.15 \\
\textbf{Elastic Horizon} & \textbf{39.47 $\pm$ 2.38} & \textbf{71.62 $\pm$ 1.96} \\
\bottomrule
\end{tabular}
\end{table}

\section{Trajectory Length Distribution Analysis}
\label{app:distribution}



\begin{table}[t]
\centering
\caption{Quantile statistics of successful trajectory lengths at training step 100.}
\label{tab:quantiles}
\small
\begin{tabular}{lccccc}
\toprule
\textbf{Benchmark} & \textbf{Min}& \textbf{Mean}  & \textbf{P50} & \textbf{P90} & \textbf{Max} \\
\midrule
AppWorld & 9 & $18.14$  & $14$ &  $37.21$ &  $50$ \\
BFCL     & 1 & $11.93$  & $11$  & $24.29$ & $50$\\
\bottomrule
\end{tabular}
\end{table}

\paragraph{Observations supporting P90.}
\begin{enumerate}[leftmargin=*, itemsep=2pt]
    \item \textbf{Right skew}: Both distributions have positive skewness, confirming that trajectory lengths are heavy-tailed rather than symmetric.
    
    \item \textbf{Mean underestimates capability}: The mean is substantially lower than P90 , indicating that using the mean would set an overly conservative horizon that fails to accommodate harder task instances.
    
    \item \textbf{Max is unstable}: The gap between P90 and Max reflects outlier trajectories from inefficient explorations. Using Max would inflate the horizon estimate and destabilize training.
    
    \item \textbf{P90 balances coverage and robustness}: P90 covers 90\% of successful trajectories while excluding the top 10\% of potential outliers, providing a stable yet ambitious capability estimate.
\end{enumerate}

\section{Additional Ablation Results}
\label{app:ablation}

\subsection{Full Percentile Comparison}

To answer ``why P90 over other percentiles?'', we report two complementary views: (i) the success rate achieved on each benchmark when the controller uses different statistics of $\mathcal{B}$ (Table~\ref{tab:ablation_percentile_cross}); and (ii) the converged horizon, its stability, and the resulting interpretation on AppWorld (Table~\ref{tab:ablation_percentile_full}).

\begin{table}[t]
\centering
\caption{Percentile choice across benchmarks (Mean@8, \%). P90 is the best statistic on both benchmarks.}
\label{tab:ablation_percentile_cross}
\small
\begin{tabular}{lcc}
\toprule
\textbf{Statistic} & \textbf{AppWorld SR (\%)} & \textbf{BFCL SR (\%)} \\
\midrule
P50          & 24.12          & 68.37 \\
\textbf{P90} & \textbf{39.47} & \textbf{71.62} \\
P95          & 38.38          & 69.87 \\
Max          & 36.0           & 62.88 \\
\bottomrule
\end{tabular}
\end{table}

Performance increases monotonically from P50 to P90 on both benchmarks, then gradually decreases toward Max, consistent with the heavy-tail interpretation in Appendix~\ref{app:theory}.

\begin{table}[t]
\centering
\caption{Detailed percentile ablation on AppWorld (Step 200). We report success rate (SR), final converged horizon, and horizon stability (standard deviation over the last 50 steps).}
\label{tab:ablation_percentile_full}
\footnotesize
\begin{tabular}{lccc}
\toprule
\textbf{Statistic} & \textbf{SR (\%)} & \textbf{Final $K$} & \textbf{$K$ Std} \\
\midrule
Mean         & 24.12          & 19.09          & 16.4          \\
P50          & 24.12          & 15.82          & 14.9          \\
\textbf{P90} & \textbf{39.47} & 28.1          & 12.8          \\
P95          & 38.38          & 44.15          & 9.7           \\
Max          & 36.0           & 50.00          & 3.1           \\
\bottomrule
\end{tabular}
\end{table}

\paragraph{Analysis.}
Lower percentiles (Mean, P50) cause premature horizon convergence ($K \approx 15\text{-}19$), limiting the agent's ability to solve harder tasks that require longer interactions. Higher percentiles (P95, Max) are sensitive to outlier trajectories, pushing $K$ toward $K_{\max}$ and losing the efficiency benefit of adaptive control. P90 provides the best tradeoff: sufficiently high to accommodate challenging tasks, yet robust to occasional outliers.

\subsection{Headroom Sensitivity}

Table~\ref{tab:ablation_headroom_full} provides complete results for the headroom ablation.

\begin{table}[t]
\centering
\caption{Ablation on headroom $\Delta$ (AppWorld, Step 200).}
\label{tab:ablation_headroom_full}
\small
\begin{tabular}{lccc}
\toprule
\textbf{$\Delta$} & \textbf{SR (\%)} & \textbf{Final $K$} & \textbf{Tokens} \\
\midrule
0  & 15.35 & 6.8  & $2.2\times10^{5}$ \\
5  & 12.06 & 22.9 & $3.7\times10^{5}$ \\
10 & 39.47 & 28.1 & $3.9\times10^{5}$ \\
15 & 25.87 & 28.7 & $3.5\times10^{5}$ \\
20 & 33.11 & 35.9 & $3.7\times10^{5}$ \\
\bottomrule
\end{tabular}
\end{table}

\paragraph{Analysis.}
Setting $\Delta = 0$ removes the exploration margin entirely: the target horizon can never exceed the current capability estimate, so the horizon collapses toward $K_{\min}$ ($K = 6.8$) and the agent never attempts tasks longer than what it already solves. Beyond this degenerate case, success rate is not monotone in $\Delta$, and the gaps between neighbouring settings are comparable to the seed-to-seed spread reported in Table~\ref{tab:main_results_std}; we therefore claim only that $\Delta = 10$ is the best of the values we tried, not that a smooth trend exists. Larger headroom does raise the converged horizon ($K = 35.9$ at $\Delta = 20$), moving the controller toward fixed long-horizon behavior and eroding the efficiency benefit of adaptive control.

\subsection{EMA Coefficient Sensitivity}

\begin{table}[t]
\centering
\caption{Ablation on EMA coefficient $\alpha$ (AppWorld).}
\label{tab:ablation_ema_full}
\small
\setlength{\tabcolsep}{4pt}
\begin{tabular}{lcccl}
\toprule
\textbf{$\alpha$} & \textbf{SR (\%)}& \textbf{Final $K$} & \textbf{$K$ Std} & \textbf{Behavior} \\
\midrule
0.1 & 39.47 & 28.1 & 12.8 & Slow, stable \\
0.2 & 30.92 & 27.6 & 14.1 & Mild oscillation \\
1.0 & 24.78 & 28   & 17.9 & Unstable \\
\bottomrule
\end{tabular}
\end{table}

\paragraph{Analysis.}
The EMA coefficient $\alpha$ controls the tradeoff between responsiveness and stability. Small $\alpha$ (e.g., 0.1) yields very smooth trajectories but slow adaptation to capability changes. Large $\alpha$ (e.g., 1.0, equivalent to no smoothing) tracks the raw P90 estimate directly, causing oscillations due to sampling variance. 

\subsection{Sample Efficiency Comparison}
\label{app:efficiency}

Table~\ref{tab:efficiency_appworld} reports cumulative trajectory tokens consumed by each method on AppWorld at two operating points: (i) the first training step at which success rate reaches 50\%, and (ii) the step at which each method attains its peak success rate. The ``Token Savings'' column is computed against GRPO ($K{=}50$) at the matched SR=50\% operating point.

\begin{table*}[t]
\centering
\caption{Sample efficiency on AppWorld (Qwen2.5-7B). Tokens are cumulative trajectory tokens (M = millions). Peak SR is the maximum Mean@8 attained during training. ``$-$'' indicates the method did not reach the SR threshold. Token savings are computed against GRPO ($K{=}50$) at the matched SR=50\% operating point.}
\label{tab:efficiency_appworld}
\small
\setlength{\tabcolsep}{6pt}
\begin{tabular}{lcccc}
\toprule
\textbf{Method} & \textbf{Tok @ SR=50\%} & \textbf{Tok @ Peak} & \textbf{Peak SR (\%)} & \textbf{Token Savings} \\
\midrule
GRPO ($K{=}50$)         & 50.21M          & 90.85M          & 56.49          & (baseline) \\
ScalingInter (Linear)   & 73.98M          & 73.98M          & 50.44          & --- \\
TTI (Multiplicative)    & ---             & 85.07M          & 49.50          & --- \\
\textbf{Elastic Horizon} & \textbf{36.28M} & \textbf{85.31M} & \textbf{56.87} & \textbf{$\sim$28\%} \\
\bottomrule
\end{tabular}
\end{table*}

\begin{figure}[t]
    \centering
    \includegraphics[width=0.85\columnwidth]{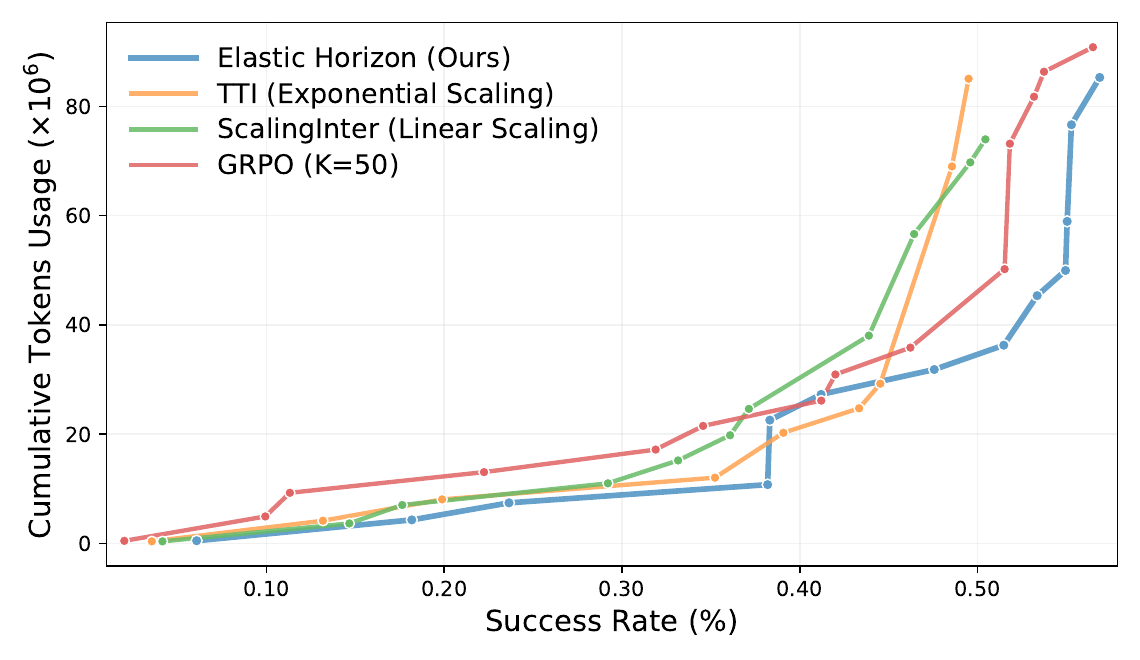}
    \caption{
        \textbf{Success rate vs.\ cumulative trajectory tokens (AppWorld, Qwen2.5-7B).}
        Elastic Horizon reaches matched success rates with strictly fewer cumulative tokens than every fixed-horizon baseline. At SR${=}50\%$, the cumulative gap is 28\% vs.\ GRPO ($K{=}50$).
    }
    \label{fig:efficiency_cumulative}
\end{figure}

\paragraph{Analysis.}
Elastic Horizon reaches SR=50\% with only 36.28M cumulative tokens, compared to 50.21M for GRPO ($K{=}50$), a \textbf{28\% reduction at equivalent performance}. At peak performance, Elastic Horizon achieves the highest SR (56.87\%) while consuming fewer total tokens than the strongest fixed-horizon baseline. Figure~\ref{fig:efficiency_cumulative} visualizes the underlying trajectories: Elastic Horizon (blue) reaches matched SR with strictly fewer cumulative tokens than every fixed-horizon baseline. The 28\% saving stems from two sources: shorter trajectories during early training (when the controller has not yet expanded to its converged value), and automatic convergence that avoids wasteful exploration beyond $H^*$. These two contributions correspond to the per-step gap visualized in Figure~\ref{fig:efficiency} of the main text.

\begin{figure}[t]
    \centering
    \includegraphics[width=0.7\columnwidth]{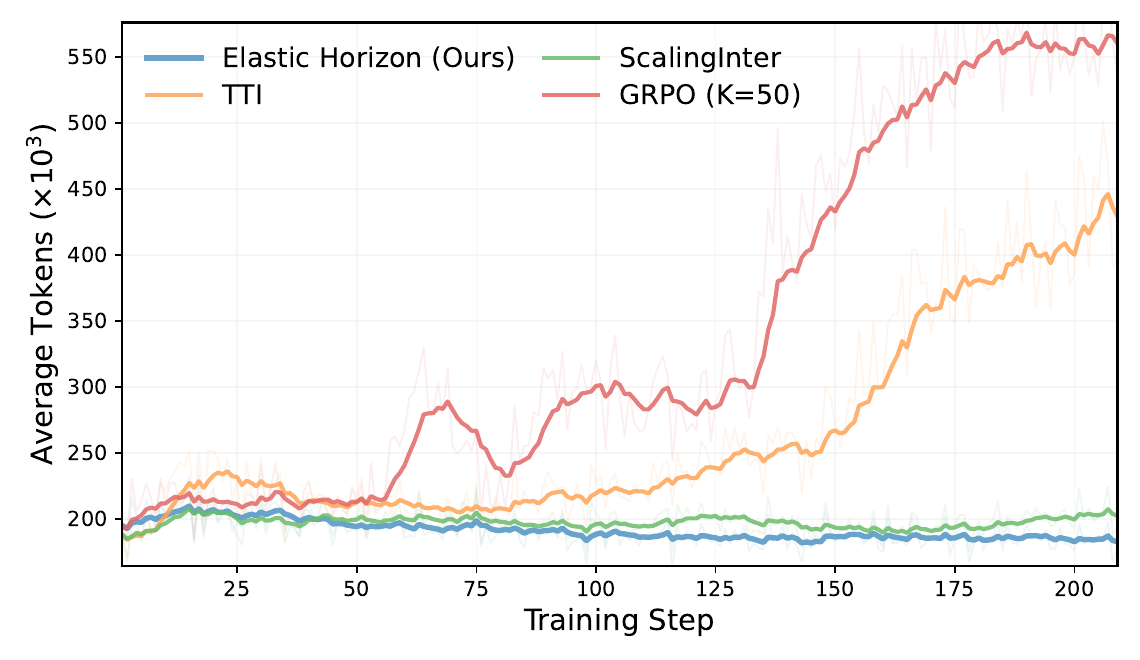}
    \caption{
        \textbf{Average tokens per training step on BFCL 7B.}
        Elastic Horizon (blue) consumes fewer per-step tokens than open-loop baselines throughout training, with the gap most visible during the rapid expansion phase of TTI/ScalingInter.
    }
    \label{fig:efficiency_bfcl}
\end{figure}

\paragraph{BFCL average tokens per step.}
Figure~\ref{fig:efficiency_bfcl} reports per-step token consumption on BFCL 7B for completeness. The trend mirrors AppWorld 14B: Elastic Horizon maintains lower per-step trajectory tokens than open-loop schedules throughout training. We move this panel to the appendix because the main-text efficiency story (Figure~\ref{fig:efficiency}) is built around AppWorld 14B, where per-step gains are most pronounced.

\section{Limitations and Future Directions}
\label{app:limitations}

\subsection{Cold Start Phase}

During early training when successful trajectories are scarce, the P90 estimate may be unreliable. Our current solution (maintaining the initial horizon until buffer fills) is conservative but may delay adaptation. Future work could explore:
\begin{itemize}[leftmargin=*, itemsep=2pt]
    \item Using trajectory lengths from \emph{all} trajectories (not just successful ones) with appropriate weighting.
    \item Initializing the buffer with trajectories from a pretrained policy.
    \item Hybrid strategies that use open-loop schedules during cold start, then switch to closed-loop.
\end{itemize}

\subsection{Single Global Horizon}

Elastic Horizon adjusts a single global horizon $K_t$ applied to all tasks. For task distributions with high variance in complexity, a single horizon may be suboptimal: easy tasks waste budget, while hard tasks may be truncated. Future work could explore:
\begin{itemize}[leftmargin=*, itemsep=2pt]
    \item Per-task or per-difficulty-level horizon allocation.
    \item State-conditioned horizon prediction.
    \item Integration with task curriculum methods that also adapt task selection.
\end{itemize}

\subsection{Non-Stationary Task Distributions}

Our convergence analysis assumes the task distribution is stationary. If the distribution shifts during training (e.g., curriculum over tasks), the P90 estimate may lag. The FIFO buffer provides some adaptivity, but stronger distribution shift may require:
\begin{itemize}[leftmargin=*, itemsep=2pt]
    \item Drift detection mechanisms.
    \item Weighted buffers that discount older samples.
    \item Explicit handling of task difficulty metadata.
\end{itemize}

\subsection{Theoretical Gaps}

The formal discussion in Appendix~\ref{app:theory} is intentionally lightweight: we do not establish a closed-form bound between $\hat{B}_t$ and $H^*$, and we analyze the EMA update only under the simplifying assumption of a stationary target. Several theoretical questions remain open:
\begin{itemize}[leftmargin=*, itemsep=2pt]
    \item Under what conditions is P90 \emph{optimal} among percentile-based estimators?
    \item Can we derive the headroom $\Delta$ from first principles rather than tuning?
    \item How does the effective frontier $H^*$ evolve during training under the joint policy--controller dynamics, and can this be characterized formally?
\end{itemize}

These questions present opportunities for future theoretical investigation.

\section{Extended Discussion}
\label{app:discussion}

This appendix elaborates on two aspects of Elastic Horizon that are summarized only briefly in the main Conclusion.

\paragraph{What the frontier captures.}
The frontier $H^*$ identified by Elastic Horizon reflects \emph{computational complexity}---the number of interaction steps an agent needs to solve tasks---rather than \emph{conceptual difficulty}. A task may be conceptually simple yet require many API calls (e.g., iterating over a list), or conceptually hard yet solvable in few steps (e.g., a single complex reasoning query). Elastic Horizon optimizes the former dimension because it directly determines training compute, and it does so without requiring practitioners to specify growth rates, stage boundaries, or maximum horizons, all of which depend on unknown task properties. This self-regulation is particularly valuable when deploying RL to new environments with unknown task complexity, when the task distribution is heterogeneous, and whenever compute is the binding constraint.

\paragraph{Complementarity with test-time horizon control.}
Elastic Horizon operates during training to optimize sample efficiency, while concurrent test-time methods such as BATS~\citep{liu2025bats} operate during deployment to optimize task performance under budget constraints. The two approaches address different stages of the agent lifecycle and are naturally complementary: combining training-time horizon adaptation with test-time budget awareness is a direction for end-to-end efficient agent systems.

\end{document}